\documentclass{article}

\usepackage{arxiv}

\usepackage[utf8]{inputenc} 
\usepackage[T1]{fontenc}    
\usepackage{hyperref}       
\usepackage{url}            
\usepackage{booktabs}       
\usepackage{amsfonts}       
\usepackage{nicefrac}       
\usepackage{microtype}      
\usepackage{lipsum}		
\usepackage{amsmath}
\usepackage{graphicx}
\usepackage[ruled,vlined]{algorithm2e}

\renewcommand{\vec}[1]{\boldsymbol{#1}}     

\newcommand{\mat}[1]{\mathbf{#1}}           

\newcommand{\X}{\mathcal{X}}

\newcommand{\R}{\mathbb{R}}
\DeclareMathOperator{\E}{E}
\newcommand{\Pro}{\mathrm{P}}

\usepackage{fancyhdr}

\title{Missing Features Reconstruction Using a Wasserstein Generative Adversarial Imputation Network}
\date{} 					

\author{
  Magda Friedjungová \\
  Faculty of Information Technology\\
  Czech Technical University in Prague\\
  Prague, Czech Republic\\
  \texttt{magda.friedjungova@fit.cvut.cz} \\
   \And
  Daniel Vašata \\
  Faculty of Information Technology\\
  Czech Technical University in Prague\\
  Prague, Czech Republic\\
  \texttt{daniel.vasata@fit.cvut.cz} \\
  \And
  Maksym Balatsko \\
  Faculty of Information Technology\\
  Czech Technical University in Prague\\
  Prague, Czech Republic\\
  \texttt{balatmak@fit.cvut.cz} \\
  \And
  Marcel Jiřina \\
  Faculty of Information Technology\\
  Czech Technical University in Prague\\
  Prague, Czech Republic\\
  \texttt{marcel.jirina@fit.cvut.cz} \\
}

\renewcommand{\undertitle}{A preprint of paper accepted to ICCS 2020.}

\begin{document}
\maketitle

\begin{abstract}
Missing data is one of the most common preprocessing problems.
In this paper, we experimentally research the use of generative
and non-generative models for feature reconstruction.
Variational Autoencoder with Arbitrary Conditioning (VAEAC)
and Generative Adversarial Imputation Network (GAIN) were researched as representatives of
generative models, while the denoising autoencoder (DAE) represented non-generative models.
Performance of the models is compared to traditional methods $k$-nearest neighbors ($k$-NN) and
Multiple Imputation by Chained Equations (MICE).
Moreover, we introduce WGAIN as the Wasserstein modification of GAIN,
which turns out to be the best imputation model when the degree of missingness
is less than or equal to $30\%$.
Experiments were performed on real-world and artificial datasets with continuous features
where different percentages of features, varying from
$10\%$ to $50\%$, were missing. Evaluation of algorithms was done by measuring
the accuracy of the classification model previously trained on the
uncorrupted dataset.
The results show that GAIN and especially WGAIN are the best imputers regardless of the conditions.
In general, they outperform or are comparative to MICE, $k$-NN, DAE, and VAEAC.
\end{abstract}

\keywords{Imputation Methods \and Feature Reconstruction \and Missing Data \and Generative Models \and Autoencoders \and Wasserstein GAN}

\section{Introduction}
\label{sec:intro}
When working with real-world datasets one of the standard problems that needs solving as part of the
data preprocessing phase is dealing with missing data.
The missingness can be represented by either individual missing data randomly located in instances
or by the absence of entire features.

To our best knowledge, not much attention is paid to the second scenario where
entire features are missing, i.e., there are no clear answers to questions such as how to face the situation,
how the standard imputation method will perform or if there is a need to approach this challenge
in a different way.

The aim of our work is to study these issues by experimentally comparing several state-of-the
art imputation methods in real-world scenarios where one needs to impute (i.e., reconstruct) entire features.
This work follows up on our previous work presented in paper~\cite{iccs2019}, where we focus on the comparison of traditional
($k$-NN, linear regression, MICE) and modern (multi-layer perceptron, extreme gradient boosted trees) imputation methods.

In the current paper, we research more universal imputers represented by autoencoders and generative neural network models.
These models have a common advantage in that one does not need to know which features are missing in advance.
On the contrary, regular imputation methods need to be trained for each combination of missing features separately.
A typical example where a universal imputer is needed is the prediction of a classification model
from sensor data, where a sensor breakdown leads to missing data in one or more features.
Usually, the prediction model itself is not able to handle this situation without a significant decrease
in its performance.
Furthermore, one typically does not know in advance which sensor is going to be broken.
The best approach would be to retrain the model using data without missing features.
However, in a production setting model retraining is
impossible as the existing model needs to respond to corrupted data immediately.

We consider a situation where the prediction model is trained on a
complete preprocessed dataset with numeric features, and we study its accuracy changes on new unseen data with imputed missing features.
The amount of missing data (i.e. features) varies between $10\%$ and $50\%$.
Experiments are performed on ten real and two artificial datasets.
The impact of imputation is measured as the classification accuracy change of the best performing from six commonly used
classification models: logistic regression, multi-layer perceptron, $k$-NN, naive Bayes,
extreme gradient boosted trees~\cite{chen16}, and random forest.
Besides accuracy we also use root mean squared error (RMSE) (which was also used in~\cite{GAIN,Camino2019,VAEAC})
as a measure of the quality of the imputation.

We compare the denoising autoencoder (DAE)~\cite{dae},
Generative Adversarial Imputation Network (GAIN)~\cite{GAIN}, and Variational Autoencoder with Arbitrary Conditioning (VAEAC)~\cite{VAEAC}
with $k$-NN and MICE~\cite{mice}, which are considered to be successful traditional imputation methods.
Moreover, we introduce Wasserstein Generative Adversarial Imputation Network (WGAIN), a Wasserstein based modification of GAIN,
see~\cite{WassersteinGAN}. WGAIN is a generative imputation model and generally outperforms
other presented models on the tested datasets.
The Earth-Mover distance and the corresponding discriminator's critic of the Wasserstein approach do not suffer
from vanishing gradients in the way that a vanilla GAN would. This enables the model to capture the desired distribution better.

The paper is organized as follows. In Section~\ref{sec:related}, we briefly review related work in this field.
In Section~\ref{sec:model} the WGAIN model is introduced. Section~\ref{sec:experiments} is devoted
to the description of experiments performed, including the evaluation of their results.
Finally, the paper is concluded in Section~\ref{sec:conclusion}.

\section{Related Work}
\label{sec:related}
There are many traditional imputation methods, such as e.g.,~\cite{alireza08,little2014,buuren2018}.
Some of the most common and successful are $k$-nearest neighbors imputation
($k$-NN)~\cite{Jonsson2004} and multivariate imputation by chained equations (MICE)~\cite{Schafer1997,buuren2018}.

Approaches based on deep learning have been under active development for the last few years.
They use many variants of neural networks starting from
multi-layer perceptron, e.g., in~\cite{Ramirez2015,Wozniak2018}.
A more advanced approach is based on the autoencoder as a specific kind of neural network
aiming to reconstruct inputs on its outputs. Here, one of the
most commonly used models is the denoising autoencoder (DAE)~\cite{dae}, e.g.,~\cite{Wong2014,DUAN2016,Moore2017,mida,Costa2018}.
Typically, they are used in a discriminative way (see~\cite{mida} for difference),
meaning they impute a single value, which is deterministic once the network is trained.

On the other hand, the most recent research focuses on generative models which enables one
to sample from the distribution conditioned on the observed features and thus get information about the uncertainty in imputed values.
There are two groups of deep learning generative models. First, there are models based on
the variational autoencoder (VAE)~\cite{VAE} and its conditional alternations, see~\cite{Sohn2015,Mccoy2018,VAD,Lopez-Martin2017}.
In this group, some of the most successful imputation models are VAEAC~\cite{VAEAC} and HI-VAE~\cite{HIVAE}.

The second group contains models based on the Generative Adversarial Network (GAN)~\cite{GAN}.
Notably, one can encounter them in image reconstruction tasks (i.e., image inpainting), see~\cite{collagan,misgan,Pathak2016}.
One of the most prominent methods based on GAN is the GAIN~\cite{GAIN}, which uses the generator discriminator mechanism
to achieve learning of the desired distribution.
The generator observes some components of a real data vector, imputes the missing components conditioned on what is
observed, and outputs a completed vector. The discriminator then takes a completed vector
and attempts to determine which components were observed and which were imputed.
The GAIN forms the base for our modification of the imputation method based on Wasserstein GAN~\cite{WassersteinGAN}, which is introduced in the next section.
Only recently, GAIN was outperformed by the previously mentioned VAEAC and HI-VAE.
However, for numeric variables, HI-VAE achieves a comparable error to the rest of the methods~\cite{HIVAE}.
Therefore we have chosen only VAEAC for the experimental comparison.

\section{Wasserstein Generative Adversarial Imputation Network}
\label{sec:model}
In this section, the WGAIN model is introduced as GAIN adapting
the discriminative approach from Wasserstein GAN.

Let us denote $\X = \R^d$ the $d$-dimensional numeric data domain and let $\vec X = (X_1,\dotsc, X_d)$
be a random vector with values in $\X$
whose distribution is denoted by $\Pro(\vec X)$. Let the mask be a random binary vector $\vec M$,
i.e., random vector with values in $\{0,1\}^d$. The mask corresponds to unobserved values of $\vec X$ so that
the value $0$ of its $j$th component means that the $j$th feature of $X_j$ is missing
and the value $1$ means that the $j$th feature of $X_j$ is not missing.
The distribution of $\vec M$ corresponds to the distribution of missingness in the data.
Let us further denote by $\tilde{\vec X}$ the vector $\vec X$ having zeros in place of missing
values given by
\[
  \tilde{\vec X} = \vec X \odot \vec M,
\]
where $\odot$ denotes element-wise multiplication.
Our aim is to impute missing values in $\tilde{\vec X}$ based on information from non-missing features
of $\tilde{\vec X}$ and $\vec M$.
It is done in a generative way and it means that we want to learn the conditional distribution
$\Pro(\vec X | \tilde{\vec X} = \tilde{\vec x}, \vec M = \vec m)$ of $\vec X$ given
$\tilde{\vec X} = \tilde{\vec x}$ and $\vec M = \vec m$.
To do this let $\vec Z$ be a random vector with identically distributed independent components having normal
distribution $\text{N}(0,\sigma^2)$ with variance $\sigma^2$ and define
\[
  \tilde{\vec X}_{\vec Z} = \vec Z \odot (1 - \vec M) + \vec X \odot \vec M,
\]
i.e. $\tilde{\vec X}_{\vec Z}$ is $\tilde{\vec X}$ with missing components replaced by normal random variables.

\begin{figure}
  \centering
  \includegraphics[width=0.9\textwidth]{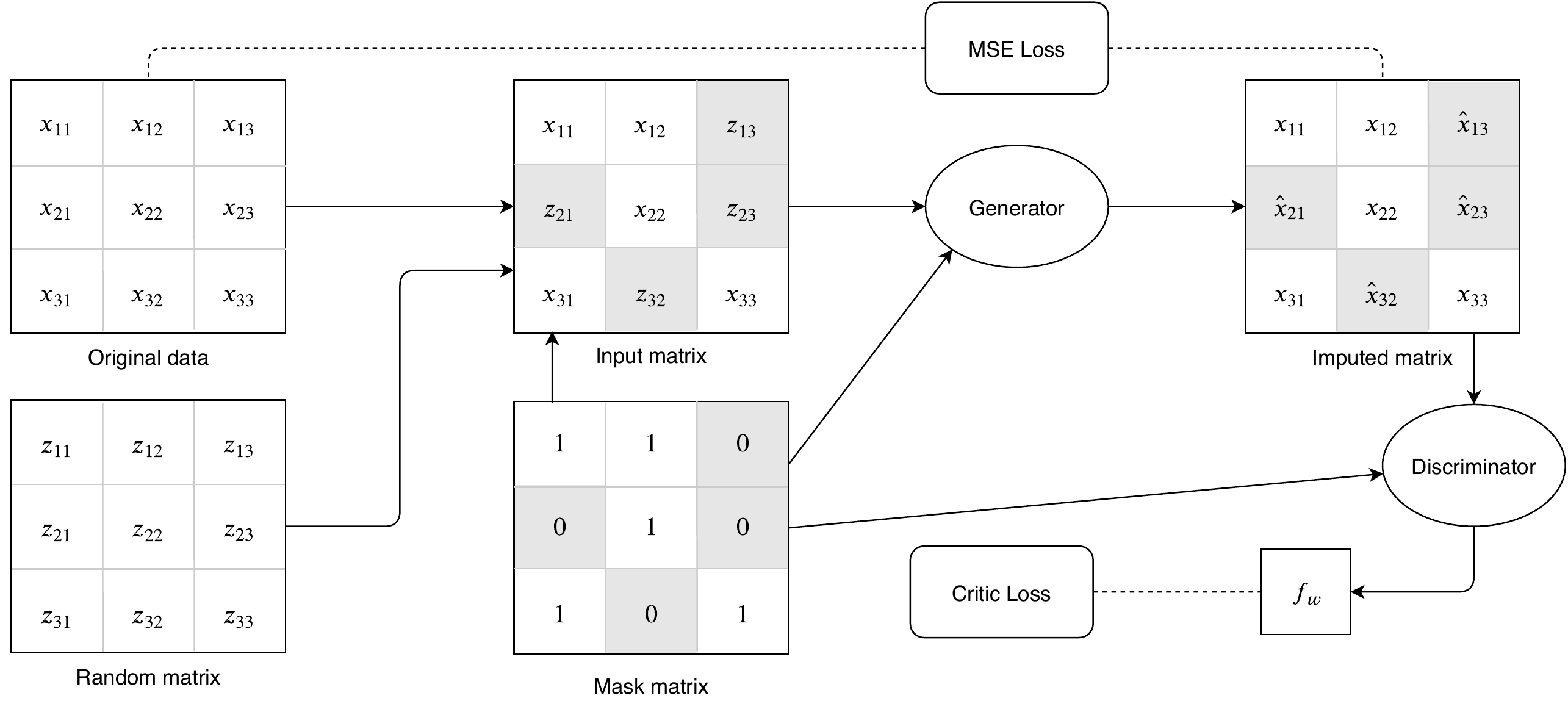}
  \caption{WGAIN structure and mini-batch data flow.}
  \label{fig:wgain_arch}
\end{figure}
The WGAIN model consists of two parts, the generator $g$ and the critic $f$, both represented by deep neural networks.
The generator $g$ is constructed as a mapping $g: \X \times \{0,1\}^d \to \X$ so that
\[
  \hat{\vec X}_{\vec Z} = g(\tilde{\vec x}_{\vec Z}, \vec m) \odot (1 - \vec m) + \tilde{\vec x} \odot \vec m
\]
is a random vector whose conditional distribution $\Pro(\hat{\vec X}_{\vec Z}| \tilde{\vec X} = \tilde{\vec x}, \vec M = \vec m)$, determined by the distribution $\Pro(\vec Z)$ of $\vec Z,$
should be close to the conditional distribution $\Pro(\vec X | \tilde{\vec X} = \tilde{\vec x}, \vec M = \vec m)$.
Note that $g(\tilde{\vec x}_{\vec Z}, \vec m)$ is a random vector corresponding to $\tilde{\vec x}$ with all missing
components imputed.

In order to train it, we employ the standard squared loss function
\[
  L_{\text{MSE}}(\hat{\vec x}_{\vec z}, \vec x) = \lVert\hat{\vec x}_{\vec z} - \vec x\rVert^2,
\]
forcing the output $\hat{\vec X}_{\vec Z}$ to be close to the original data $\vec X$.
However, it turns out that this condition alone is not sufficient for learning the proper conditional distribution.
To improve the performance of the generator, one may introduce a discriminator trying to find out which
components of $\hat{\vec X}_{\vec Z}$ were imputed and use the discriminator for adversarial training.
This approach was introduced in~\cite{GAIN} and is the base of WGAIN.

In this paper we present a similar way how to improve the conditional distribution of the generator's output.
It is based on the Earth-Mover (EM) distance between two probability distributions $\Pro(X), \Pro(Y)$
defined by
\[
  W\big(\Pro(X), \Pro(Y)\big) =
  \inf_{\gamma \in \mat\Pi(\Pro(X), \Pro(Y))} \E_{(X, Y) \sim \gamma} \lVert X - Y \rVert,
\]
where $\mat\Pi(\Pro(X), \Pro(Y))$ denotes the set of all joint distributions
$(X, Y)$ whose marginals are respectively $\Pro(X)$ and $\Pro(Y)$.
The term $\E_{(X, Y) \sim \gamma} \lVert X - Y \rVert$ might be understood as a measure
of how much probability mass has to be transported in order to transform the distributions $\Pro(X)$
into the distribution $\Pro(Y)$ when the joint distribution is $\gamma$.
The EM distance can thus be seen as the cost of the optimal transport plan, see~\cite{WassersteinGAN}
and references therein for more details.
The EM distance is usually expressed using the Kantorovich-Rubinstein duality as
\begin{equation}
  \label{eq:EM_distance}
  W\big(\Pro(X), \Pro(Y)\big)
  = \sup_{\lVert f \rVert_L \leq 1} \E_{X \sim \Pro(X)} f(X) - \E_{Y \sim \Pro(Y)} f(Y),
\end{equation}
where $\lVert f \rVert_L$ means that $f$ is Lipschitz continuous with Lipschitz constant $1$
which might be changed to any constant $K$ since it just multiplies
$W\big(\Pro(X), \Pro(Y)\big)$ by the same constant.

In Wasserstein GAN one approximates \eqref{eq:EM_distance} by training the neural network
$f_{\vec w}$ parametrized with weights $\vec w$ in some compact space $\mathcal{W}$,
thus enforcing the Lipschitz continuity. The function $f_{\vec w}$ is called the \emph{critic} and is trained
to maximize the expectations difference in \eqref{eq:EM_distance}. For a single dimensional generator
$g$ trying to transform random variable $Z$ so that it has the distribution $\Pro(X)$ one maximizes
\[
  \max_{\vec w \in \mathcal{W}} \E_{X \sim \Pro(X)} f_{\vec w}(X)
  - \E_{Z \sim \Pro(Z)} f_{\vec w}(g(Z)).
\]

In our case we want to minimize the EM distance between
$\Pro(\hat{\vec X}_{\vec Z}| \tilde{\vec X} = \tilde{\vec x}, \vec M = \vec m)$ and $\Pro(\vec X | \tilde{\vec X} = \tilde{\vec x}, \vec M = \vec m)$.
Hence, we take the mask $\vec M$ as the second argument of the critic as additional information to the first argument
given by $\vec X$ with correct features behind the mask $\vec M$.
The critic is therefore a mapping $f_{\vec w}: \X \times \{0,1\}^d \to \R$ trained to maximize
\[
  \max_{\vec w \in \mathcal{W}} \E_{\vec X \sim \Pro(\vec X)} f_{\vec w}(\vec X, \vec M)
  - \E_{\vec Z \sim \Pro(\vec Z)} f_{\vec w}(\hat{\vec X}_{\vec Z}, \vec M),
\]
which is usually estimated by sample means from mini-batches. The overall structure of WGAIN is depicted in Figure \ref{fig:wgain_arch}.

\subsection{Training}
\label{sec:training}
The critic $f_{\vec w}$ is used in adversarial training of both the generator $g$ and the critic itself.
There the generator and the critic play an iterative two-player minimax game
when the critic wants to recognize the imputed values from the real ones and
the goal of the generator is to trick the critic so it cannot recognize them.
Moreover, the generator's output is tighten to the correct output by the squared loss function $L_{\text{MSE}}$.

Putting it all together, we have two objective functions to minimize. The first corresponds to training of the
discriminator given by
\[
  J(f_{\vec w}) = \lambda_{f_{\vec w}} \Big(\E_{\vec Z \sim \Pro(\vec Z)} f_{\vec w}(\hat{\vec X}_{\vec Z}, \vec M) - \E_{\vec X \sim \Pro(\vec X)} f_{\vec w}(\vec X, \vec M)\Big),
\]
where the weight $\lambda$ enables one to increase or decrease the influence of the corresponding gradient.
Second is the objective for the generator,
\[
  J(g) = - \lambda_g \E_{\vec Z \sim \Pro(\vec Z)} f_{\vec w}(\hat{\vec X}_{\vec Z}, \vec M)
  + \lambda_{\text{MSE}} \E_{\vec X \sim \Pro(\vec X), \vec Z \sim \Pro(\vec Z)} L_{\text{MSE}}(\hat{\vec X}_{\vec Z}, \vec X),
\]
where the first term  $\lambda_g$ and $\lambda_{\text{MSE}}$ are weights enabling one to strengthen or weaken the influence of squared loss function.
The optimization is done via alternating gradient descent, where the
first step is updating the critic $f_{\vec w}$ and the second step is
updating the generator $g$. Hence, when perfectly trained, the discriminator gives negative values to cases with imputed features and positive values for cases with true features. On the other hand, the generator entering the critic will be pushed to obtain large positive values of the critic as it gives to real values.

The pseudo-code of the WGAIN training is given in Algorithm~\ref{alg:WGAIN}.

\begin{algorithm}[t]
\SetAlgoLined
\KwIn{$\alpha$ - the learning rate; $w_{\max}$ - maximal norm used in clipping; $m$ - the mini-batch size
}
 Draw $m$ samples from the dataset $\{\vec x_j\}_{j=1}^{m}$\;
 Draw $m$ samples from the mask distribution $\{\vec m_j\}_{j=1}^{m}$\;
 Draw $m$ samples from the normal distribution of $\vec Z$, $\{\vec z_j\}_{j=1}^{m}$\;
 \While{not converged}{
  $\tilde{\vec x}_{\vec z_j} \leftarrow \vec z_j \odot (1 - \vec m_j) + \vec x_j \odot \vec m_j$\;
  $\hat{\vec x}_{\vec z_j} \leftarrow g(\tilde{\vec x}_{\vec z_j}, \vec m_j) \odot (1 - \vec m_j) + \vec x_j \odot \vec m_j$\;

  \vspace{1ex}
  Update weights $\vec w$ of $f_{\vec w}$ using RMSProp with learning rate $\alpha$ and gradient

  $\nabla J(f_{\vec w}) = \lambda_{f_{\vec w}}\nabla\left[\frac{1}{m}\sum_{i=1}^m f_{\vec w}\big(\hat{\vec x}_{\vec z_j}, \vec m_j\big)
   - \frac{1}{m}\sum_{i=1}^m f_{\vec w}\big(\vec x_j, \vec m_j\big)\right]$\;

  \vspace{1ex}
  Clip the norm of $\vec w$ by $\vec w_{\max}$\;
  \vspace{1ex}

  Update weights of $g$ using RMSProp with learning rate $\alpha$ and gradient

  $\nabla J(g) = \nabla\left[-\lambda_{g}\frac{1}{m}\sum_{i=1}^m f_{\vec w}\big(\hat{\vec x}_{\vec z_j}, \vec m_j\big)
    + \lambda_{\text{MSE}}\frac{1}{m}\sum_{i=1}^m \lVert \hat{\vec x}_{\vec z_j} - \vec x_j\rVert^2\right]$\;
 }
 \caption{WGAIN training pseudo-code}
 \label{alg:WGAIN}
\end{algorithm}

\section{Experiments}
\label{sec:experiments}
An experimental validation of WGAIN using ten real and two artificial publicly
available datasets is presented below. These datasets contain numeric data only and
are devoted to the classification task. Their overview, together with the corresponding best performing classification models,
is given in Table~\ref{tab:datasets}.

During the experiments, all datasets were divided as follows: $70\%$ of data was used to train all classification and
imputation models and $30\%$ was used as a test set to evaluate imputation performance.
The imputation models were trained to impute in scenarios where
randomly selected combinations of multiple features are missing.
The amount of missingness varies from $10\%$ to $50\%$ of missing features.
Finally, evaluation of the accuracy of the classification model combined with all
imputation methods is performed on the test dataset.

\subsection{Imputation Models and Their Parameters}
\label{sec:implementation}
Let us start with the presented WGAIN model. The generator and the critic architectures were the same for all datasets and are described in Table~\ref{wgainarch}. During the training, the following settings were used:
\begin{itemize}
\item The original data $\vec X$ are sampled in mini-batches of size $m = 128$.
\item The missingness is introduced using the mask $\vec M$ with the following distribution: for each training point, the portion of missingness is sampled from a uniform distribution between $0$ and maximum missing rate, which was chosen to be $0.3$. Then the binary elements of $\vec M$ were independently sampled with this portion of missingness, i.e., its item is $0$ with a probability which was previously sampled.
\item The components of random vector $\vec Z$ are i.i.d. with normal distribution having $0$ mean and standard deviation $0.01$.
\item The weights of the objectives functions $J(f_{\vec w})$ and $J(g)$ are $\lambda_{f_{\vec w}} = 10$, $\lambda_g = 2$, and $\lambda_{\text{MSE}} = 1$.
\item Maximal norm used in clipping of the critic weights is $w_{\max} = 1$.
\item We use RMSProp with learning rate $\alpha = 0.0001$ as optimizers.
\item The number of training epochs is $8000$.
\end{itemize}

\begin{table}[h!]
  \caption{Architecture details of the WGAIN. Abbreviation: FC=fully connected layer.}\label{wgainarch}
  \centering
  \begin{tabular}{|c|c|}
    \hline
    Layer & Generator \\
    \hline
     & concatenate data and mask \\
    1& FC-($1.5$ input dimension), ReLU \\
    2& FC-($1.25$ input dimension), ReLU \\
    3& FC-(input dimension), Linear \\
    \hline
    Layer & Critic\\ 
    \hline
     & concatenate data and mask \\
    1& FC-($1.5$ input dimension), ReLU  \\
    2& FC-($1.25$ input dimension), ReLU  \\
    3& FC-(1), Linear \\
    \hline
  \end{tabular}
\end{table}

The GAIN implementation follows the original paper~\cite{GAIN} and is analogous to the described WGAIN with the following differences:
\begin{itemize}
\item The generator architecture differs only in the sizes of layers, which are all equal to the input dimension.
\item The discriminator architecture is analogous to the generator architecture except for the sigmoid activation function on the last layer.
\item The binary elements of $\vec M$ are independently sampled with the common portion of missingness, which is $0.2$.
\item The hint rate used for the hint matrix is $0.9$.
\item As an optimizer, we use Adam with learning rate of $0.0001$.
\item The number of training epochs is $7000$.
\end{itemize}

In the case of DAE, we follow the structure presented in~\cite{mida}. For the hyper-parameters search, the hyperband~\cite{hyperband} algorithm was used. The typical best setup is the following: ELU as an activation function,
three layers in both the encoder and decoder parts, the size of the code is twice the input dimension,
and no regularization is used.

DAE, GAIN, and WGAIN models were implemented in the~\texttt{TensorFlow} library~\footnote{\texttt{TensorFlow} platform: \url{https://www.tensorflow.org}}.

The implementation of VAEAC was based on the
repository~\footnote{\texttt{VAEAC} implementation: \url{https://github.com/tigvarts/vaeac}}
corresponding to the original paper~\cite{VAEAC}. All hyper-parameters
stayed in the default settings.

For the MICE method (\textit{mice}), we used the~\texttt{IterativeImputer} class from the~\texttt{scikit-learn} library\footnote{\texttt{Scikit-learn} library: \url{https://scikit-learn.org}}. In the default settings, the implementation uses Bayesian ridge regression as the internal imputation model and multiple imputations are pooled by the mean.

The $k$-NN imputation (\textit{knn}) was implemented using
the~\texttt{fancyimpute} library~\footnote{\texttt{Fancyimpute} repository:~\url{https://github.com/iskandr/fancyimpute}}.
A missing value is imputed by sampling the mean of the values of its neighbors
weighted proportionally to their inverse distances.
In the case where multiple features are missing, we impute all missing values at once (per row).
For the hyper-parameter $k$ values $11, 13, 15, 17, 19, 21, 23, 25$ were tested.
The best $k$ was chosen based on the RMSE value.

\subsection{Evaluation}
\label{sec:evaluation}
The impact of imputation is evaluated using the classification accuracy changes of the best performing classification model chosen
from the six commonly used ones: logistic regression (LR), multi-layer perceptron (MLP), $k$-nearest neighbors (\mbox{$k$-NN}),
naive Bayes (NB), extreme gradient boosted trees (XGBT) (for details see~\cite{chen16}), and random forest (RF).
The best hyperparameters for each model were found using randomized search algorithm.
The accuracy of the best performing model for each dataset is shown in Table~\ref{tab:datasets}.
Furthermore, the root mean squared error (RMSE) between the original and the imputed data is also used for evaluation, e.g.,~\cite{GAIN,Camino2019,VAEAC}.

\begin{table}[h!]
\caption{Details of datasets with the corresponding best performing classification model and its accuracy on the test set.
The number of features (\# f.) does not include the target label. The \# r. stands for the number of records.}
\label{tab:datasets}
\scriptsize
\begin{center}
\begin{tabular}{|c|c|c|c|c|c|c|}
\hline
Name & Type & \# f. & \# r. & model name & accuracy \\
\hline
Cancer \cite{uci-ml} &  real & 9 & 683 & RF & 0.975  \\
EEG \cite{uci-ml} &  real & 14 & 14980 & $k$-NN & 0.952  \\
MAGIC \cite{uci-ml} &  real & 10 & 19020 & XGBT & 0.868  \\
Ozone-1 \cite{uci-ml} &  real & 72 & 1846 & $k$-NN & 0.977  \\
Ozone-8 \cite{uci-ml} &  real & 72 & 1848 & LR & 0.941  \\
QSAR \cite{uci-ml} &  real & 41 & 1055 & MLP & 0.868  \\
Shuttle \cite{uci-ml} &  real & 9 & 57998 & RF & 0.999  \\
Spambase \cite{uci-ml} &  real & 57 & 4597 & MLP & 0.940  \\
Waveform \cite{uci-ml} &  real & 21 & 5000 & LR & 0.869  \\
Yeast \cite{uci-ml} &  real & 8 & 1484 & XGBT & 0.578  \\
Ringnorm \cite{alcala11} &  art. & 20 & 7400 & NB & 0.979  \\
Twonorm \cite{alcala11} &  art. & 20 & 7400 & MLP & 0.979  \\
\hline
\end{tabular}
\end{center}
\end{table}

After all classification models were trained, and the most accurate one for each dataset was chosen,
they were combined with imputation methods.
Then, the accuracies of classification models on the imputed test dataset were measured.

Since it is not sound to compare accuracies for different datasets, we use a rank comparison.
To do so, the algorithms are ranked for each dataset separately, the best performing algorithm getting the rank of $1$,
the second-best rank $2$, etc. An example of accuracies and corresponding ranks for 10\% of
missingness is presented in Tables~\ref{tab:mean10} and~\ref{tab:ranks10}.
Even in cases when WGAIN is not the best, its performance is always comparable to the best performers.
The only exception is the EEG dataset, where $k$-NN imputation performs the best and
the WGAIN is in second place with a difference of almost two percent.

The algorithms can be compared, taking the mean over the datasets. The results can
be seen in Table~\ref{tab:ranks}.
When the degree of missingness varies from $10\%$ to $30\%$ the WGAIN performs the best.
When the degree of missingness is upwards of $30\%$ the GAIN outperforms the WGAIN.

\begin{table}[h!]
\caption{Mean ranks of the RMSE for different degrees of missingness.}
\label{tab:rmse}
\scriptsize
\begin{center}
\begin{tabular}{|c|c|c|c|c|c|c|}
\hline
& \multicolumn{5}{|c|}{Degree of missingness} \\
\cline{2-6}
Method & $10\%$ & $20\%$ & $30\%$ & $40\%$ & $50\%$ \\
\hline
 $k$-NN & 2.67 & 2.67 & 2.67 & 3.08 & 2.75 \\
 MICE & 3.17 & 3.50 & 3.33 & 3.00 & 2.91 \\
 DAE & 5.08 & 5.08 & 5.17 & 5.33 &  4.91 \\
 VAEAC & 3.25 & 3.33 & 3.42 & 3.17 & 3.50 \\
 GAIN & \textbf{2.17} & \textbf{2.08} & \textbf{2.17} & \textbf{2.08} & \textbf{2.50} \\
 WGAIN & 4.67 & 4.33 & 4.25 & 4.33 & 4.42 \\
\hline
\end{tabular}
\end{center}
\end{table}

\begin{table}[h!]
\caption{Mean of the accuracies for 10\% of missing features.}
\label{tab:mean10}
\scriptsize
\begin{center}
\begin{tabular}{|c|c|c|c|c|c|c|}
\hline
&$k$-NN & MICE & DAE & VAEAC & GAIN & WGAIN \\
\hline
Cancer & 0.9700 & 0.9744 & 0.9744 & 0.9749 & 0.9739 & \textbf{0.9755}\\
EEG & \textbf{0.9226} & 0.9046 & 0.8994 & 0.6374 & 0.9028 & 0.9052\\
MAGIC & \textbf{0.8562} & 0.8465 & 0.8459 & 0.8527 & 0.8522 & 0.8511\\
Ozone-1 & 0.9754 & 0.9763 & \textbf{0.9768} & 0.9762 & 0.9759 & 0.9763\\
Ozone-8 & 0.9404 & \textbf{0.9407} & 0.9405 & 0.9405 & 0.9406 & 0.9406\\
QSAR & 0.8608 & 0.8619 & 0.8615 & 0.8619 & 0.8609 & \textbf{0.8626}\\
Shuttle & 0.9995 & \textbf{0.9996} & 0.9945 & 0.9994 & 0.9992 & 0.9995\\
Spambase & \textbf{0.9363} & 0.9278 & 0.9307 & 0.9303 & 0.9339 & 0.9296\\
Waveform & 0.8603 & 0.8604 & 0.8585 & 0.8596 & \textbf{0.8605} & 0.8593\\
Yeast & 0.5516 & 0.5507 & 0.5533 & 0.5496 & 0.5541 & \textbf{0.5558}\\
Ringnorm & 0.9668 & 0.9671 & 0.9672 & 0.9673 & 0.9674 & \textbf{0.9680}\\
Twonorm & 0.9711 & 0.9716 & 0.9716 & 0.9716 & 0.9719 & \textbf{0.9723}\\
\hline
\end{tabular}
\end{center}
\end{table}

\begin{table}[h!]
\caption{Ranks of accuracies of the imputation methods for 10\% of missing features.}
\label{tab:ranks10}
\scriptsize
\begin{center}
\begin{tabular}{|c|c|c|c|c|c|c|}
\hline
&$k$-NN & MICE & DAE & VAEAC & GAIN & WGAIN \\
\hline
Cancer & 6 & 3.5 & 3.5 & 2 & 5 & 1\\
EEG & 1 & 3 & 5 & 6 & 4 & 2\\
MAGIC & 1 & 5 & 6 & 2 & 3 & 4\\
Ozone-1 & 6 & 2 & 1 & 4 & 5 & 3\\
Ozone-8 & 6 & 1 & 5 & 4 & 2.5 & 2.5\\
QSAR & 6 & 2.5 & 4 & 2.5 & 5 & 1\\
Shuttle & 2 & 1 & 6 & 3.5 & 5 & 3.5\\
Spambase & 1 & 6 & 3 & 4 & 2 & 5\\
Waveform & 3 & 2 & 6 & 4 & 1 & 5\\
Yeast & 4 & 5 & 3 & 6 & 2 & 1\\
Ringnorm & 6 & 5 & 4 & 3 & 2 & 1\\
Twonorm & 6 & 4 & 4 & 4 & 2 & 1\\
\hline
\end{tabular}
\end{center}
\end{table}

\begin{table}[h!]
\caption{Mean of the RMSE for 10\% of missing features.}
\label{tab:mean-rmse10}
\scriptsize
\begin{center}
\begin{tabular}{|c|c|c|c|c|c|c|}
\hline
&$k$-NN & MICE & DAE & VAEAC & GAIN & WGAIN \\
\hline
Cancer & \textbf{0.1905} & 0.1960 & 0.2219 & 0.1943 & 0.1959 & 0.2087\\
EEG & \textbf{16.4752} & 27.8197 & 29.1700 & 293.9315 & 21.8986 & 34.2722\\
MAGIC & \textbf{0.1821} & 0.2067 & 0.2072 & 0.1866 & 0.1844 & 0.1928\\
Ozone-1 & 0.1364 & \textbf{0.0826} & 0.1204 & 0.1047 & 0.1038 & 0.1051\\
Ozone-8 & 0.1549 & \textbf{0.0972} & 0.1473 & 0.1233 & 0.1230 & 0.1206\\
QSAR & \textbf{0.2356} & 0.3115 & 0.2505 & 0.2445 & 0.2376 & 0.2492\\
Shuttle & \textbf{0.0954} & 0.1022 & 0.1316 & 0.1085 & 0.1053 & 0.1097\\
Spambase & \textbf{0.2404} & 0.2723 & 0.2692 & 0.2659 & 0.2587 & 0.2731\\
Waveform & 0.2312 & 0.2304 & 0.2690 & 0.2301 & \textbf{0.2278} & 0.2429\\
Yeast & \textbf{0.3542} & 0.3610 & 0.3666 & 0.3585 & 0.3560 & 0.3631\\
Ringnorm & 0.3222 & \textbf{0.3184} & 0.3187 & 0.3191 & 0.3190 & 0.3282\\
Twonorm & 0.2967 & 0.2948 & 0.3081 & 0.2935 & \textbf{0.2918} & 0.2975\\
\hline
\end{tabular}
\end{center}
\end{table}

\begin{table}[h!]
\caption{Ranks of RMSE of the imputation methods for 10\% of missings.}
\label{tab:ranks-rmse10}
\scriptsize
\begin{center}
\begin{tabular}{|c|c|c|c|c|c|c|}
\hline
&$k$-NN & MICE & DAE & VAEAC & GAIN & WGAIN \\
\hline
Cancer & 1 & 4 & 6 & 2 & 3 & 5\\
EEG & 1 & 3 & 4 & 6 & 2 & 5\\
MAGIC & 1 & 5 & 6 & 3 & 2 & 4\\
Ozone-1 & 6 & 1 & 5 & 3 & 2 & 4\\
Ozone-8 & 6 & 1 & 5 & 4 & 3 & 2\\
QSAR & 1 & 6 & 5 & 3 & 2 & 4\\
Shuttle & 1 & 2 & 6 & 4 & 3 & 5\\
Spambase & 1 & 5 & 4 & 3 & 2 & 6\\
Waveform & 4 & 3 & 6 & 2 & 1 & 5\\
Yeast & 1 & 4 & 6 & 3 & 2 & 5\\
Ringnorm & 5 & 1 & 2 & 4 & 3 & 6\\
Twonorm & 4 & 3 & 6 & 2 & 1 & 5\\
\hline
\end{tabular}
\end{center}
\end{table}

The results of the ranking evaluation can be statistically evaluated using the Friedman
test~\cite{Friedman1937,Friedman1940} and the corresponding posthoc tests.
For more details, see~\cite{Demsar2006}.
P-values of Friedman $\chi^2_F$ and $F_F$ tests are shown in Table~\ref{tab:pvalue}.
One can see that from $20\%$ to $40\%$ of missing data the null-hypothesis,
that all methods perform the same, can be rejected at a $10\%$ significance level.
However, when the Bonferroni-Dunn post-hoc test is applied
the performance of WGAIN is significantly better than DAE only and just for $20\%$ and $30\%$ of missing data.

\begin{table}[h!]
\caption{P-values of Friedman $\chi^2_F$ and $F_F$ test.}
\label{tab:pvalue}
\scriptsize
\begin{center}
\begin{tabular}{|c|c|c|c|c|c|c|}
\hline
& \multicolumn{5}{|c|}{Degree of missingness} \\
\cline{2-6}
 & $10\%$ & $20\%$ & $30\%$ & $40\%$ & $50\%$ \\
\hline
$\chi^2_F$ test & 0.252 & 0.049 & 0.106 & 0.020 & 0.477 \\
$F_F$ test & 0.253 & 0.041 & 0.099 & 0.014 & 0.490 \\
\hline
\end{tabular}
\end{center}
\end{table}

\begin{table}[h!]
\caption{Mean ranks of the accuracy changes for different degrees of missingness.}
\label{tab:ranks}
\scriptsize
\begin{center}
\begin{tabular}{|c|c|c|c|c|c|c|}
\hline
& \multicolumn{5}{|c|}{Degree of missingness} \\
\cline{2-6}
Method & $10\%$ & $20\%$ & $30\%$ & $40\%$ & $50\%$ \\
\hline
 $k$-NN & 4.00 & 3.54 & 3.63 & 3.17 & 3.25 \\
 MICE & 3.33 & 4.04 & 3.75 & 4.21 & 3.71 \\
 DAE & 4.21 & 4.79 & 4.67 & 4.71 & 4.33 \\
 VAEAC & 3.75 & 3.13 & 3.50 & 3.59 & 3.54 \\
 GAIN & 3.21 & 2.83 & 2.83 & \textbf{2.21} & \textbf{2.79} \\
 WGAIN & \textbf{2.50} & \textbf{2.67} & \textbf{2.63} & 3.12 & 3.38 \\
\hline
\end{tabular}
\end{center}
\end{table}

The same ranking process is repeated for RMSE with results in Table~\ref{tab:rmse}.
An example of RMSE and corresponding ranks for 10\% of
missingness is presented in Tables~\ref{tab:mean-rmse10} and~\ref{tab:ranks-rmse10}.
Interestingly, the WGAIN performance is one of the worst, whereas the GAIN performs
the best. This is in contrary to the fact that the WGAIN imputes the best
from the accuracy point of view.
Hence, we can see that low RMSE, which is usually taken as a measure
of imputation quality may not lead to the desired performance on the target task.
On the other hand, the RMSE differences are relatively small as can be seen in Table~\ref{tab:mean-rmse10}.

\section{Conclusion}
\label{sec:conclusion}
We propose a Wasserstein Generative Adversarial Imputation Network as a new deep learning imputation model.
It is inspired by the GAIN. However, the discriminator is replaced by the Wasserstein critic.
It is known that the Wasserstein approach does not suffer
from vanishing gradients in the way that a vanilla GAN does. This enables the model to capture the desired distribution better.
One may assume such benefits in WGAIN as well. We experimentally showed that in the imputation performance
measured by classification accuracy, the WGAIN outperforms the other methods when the degree of missingness
is lower than or equal to $30\%$. In other cases, it is competitive. In future work, we would like to focus on the use
of WGAIN in image inpainting tasks.


\section*{Acknowledgements}
This research has been supported by SGS grant No. SGS20/213/OHK3/3T/18 and by GACR grant No. GA18-18080S.


\bibliography{arxiv}{} 
\bibliographystyle{unsrt}  

\end{document}